\documentclass[10pt,twocolumn,letterpaper]{article}

\PassOptionsToPackage{table}{xcolor}
\usepackage{wacv}              

\definecolor{wacvblue}{rgb}{0.21,0.49,0.74}
\usepackage[pagebackref,breaklinks,colorlinks,allcolors=wacvblue]{hyperref}
\usepackage{dblfloatfix}

\usepackage{amsfonts}
\usepackage{algorithmic}
\usepackage{algorithm}
\usepackage{array}
\usepackage{textcomp}
\usepackage{url}
\usepackage{verbatim}
\usepackage{wrapfig}
\usepackage{multirow}
\usepackage{capt-of}
\usepackage{pifont}

\usepackage{dsfont}
\usepackage{bbding}
\def\wacvPaperID{514} 
\def\confName{WACV}
\def\confYear{2027}

\title{STAR: A Spatial-Topology Aware Routing Framework for Generalizable 3D Scene Understanding}

\author{
Mingwei Xing\textsuperscript{*} \quad
Xinliang Wang\textsuperscript{*} \quad
Yifeng Shi\textsuperscript{$\dagger$}\\
KE Holdings Inc.\\
Beijing, China\\
{\tt\small
xingmingwei@stu.xmu.edu.cn \quad
wangxinliang@buaa.edu.cn \quad
shiyifeng@tju.edu.cn
}
}

\begin{document}

\maketitle

\begingroup
\makeatletter
\renewcommand{\thefootnote}{}
\renewcommand{\@makefntext}[1]{\noindent #1}
\footnotetext{%
\textsuperscript{*}Equal contribution. 
\textsuperscript{$\dagger$}Corresponding author.%
}
\makeatother
\endgroup
\begin{abstract}
    Constructing a unified 3D scene understanding model has long been hindered by the topological discrepancies across sensor modalities. While applying the Mixture-of-Experts (MoE) architecture is a flexible approach for multi-domain 3D understanding, we observe that conventional feature-only MoE routers may underrepresent local sampling topology under semantic supervision, making expert allocation difficult when semantic consistency coexists with geometric heterogeneity. To overcome this challenge, we propose \textbf{STAR} (Spatial-Topology Aware Routing Framework). Specifically, we introduce a multi-attribute self-supervised pre-training branch, covering topological and textural variations, to anchor cross-domain structural priors. Building upon this, we design a domain-aware expert branch with two mechanisms: Domain-Spatial-Guided Routing (DSR), which captures local topological variations from spatial context, and Entropy-controlled Dynamic Allocation (EDA), which adjusts the number of activated experts according to routing uncertainty. Together, these branches combine stable cross-domain representation learning with adaptive expert allocation. Extensive experiments across various tasks, encompassing both indoor and outdoor scenes, demonstrate the effectiveness of STAR. It achieves 80.1\% mIoU on the ScanNet validation set and 77.2\% mIoU on S3DIS, consistently improving over strong baselines. Code is available at our \href{https://xmw666.github.io/STAR/}{project page}.
\end{abstract}

\section{Introduction}
\label{sec:intro}
Recent advances in visual representation learning have enabled diverse perception systems, spanning dense visual understanding~\cite{xia2024vit,wang2025rt}, autonomous and multi-sensor perception~\cite{jinrang2023monouni,li2024monolss,chen2023transiff,kong2023dusa,jia2024ropebev,ju2021danet,shi2023open}, and multimodal understanding and tracking~\cite{lou2025llava,li2026cadtrack,li2026ragtrack}. In parallel, rapid progress in 3D reconstruction, generation, and spatial modeling~\cite{wang2026artifactworld,xing2026adaptsplat,jia2026you,jia2026panoworld,li2026pano2world,kang2025sat2realcity} has diversified the sensing modalities, acquisition pipelines, and geometric representations used to capture and model 3D scenes. While these developments provide increasingly rich and complementary sources of 3D data, they create a growing need for generalizable representations that can integrate heterogeneous data and support unified scene understanding across domains.
\begin{figure}[t]
    \centering
    \includegraphics[width=1.0\columnwidth]{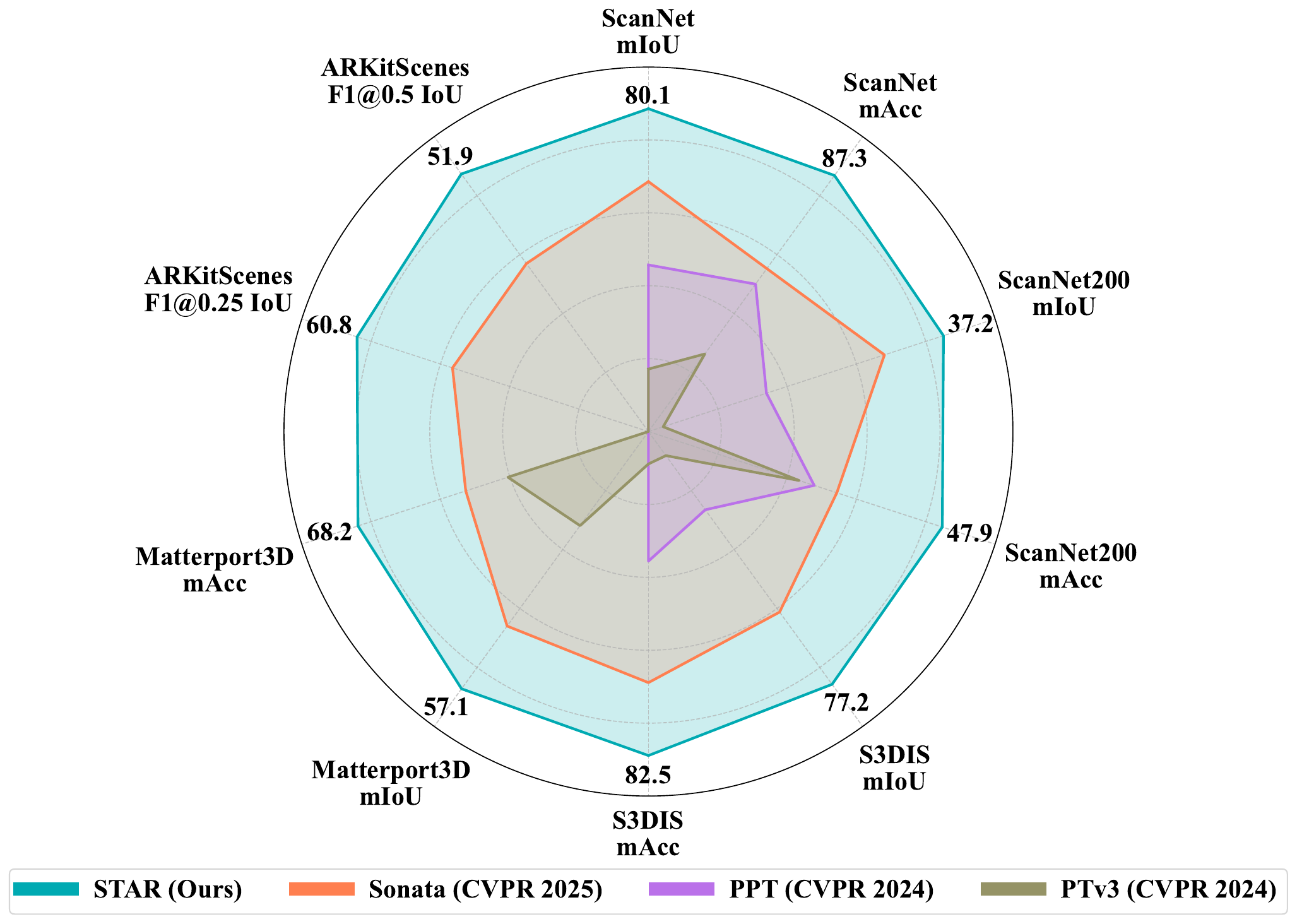}
    \caption{\textbf{Performance of STAR}. STAR achieves state-of-the-art performance across multiple 3D scene understanding benchmarks.}
    \label{fig:radar}
    \vspace{-3mm}
\end{figure}
\begin{figure*}[t]
    \centering
    \includegraphics[width=0.8\linewidth]{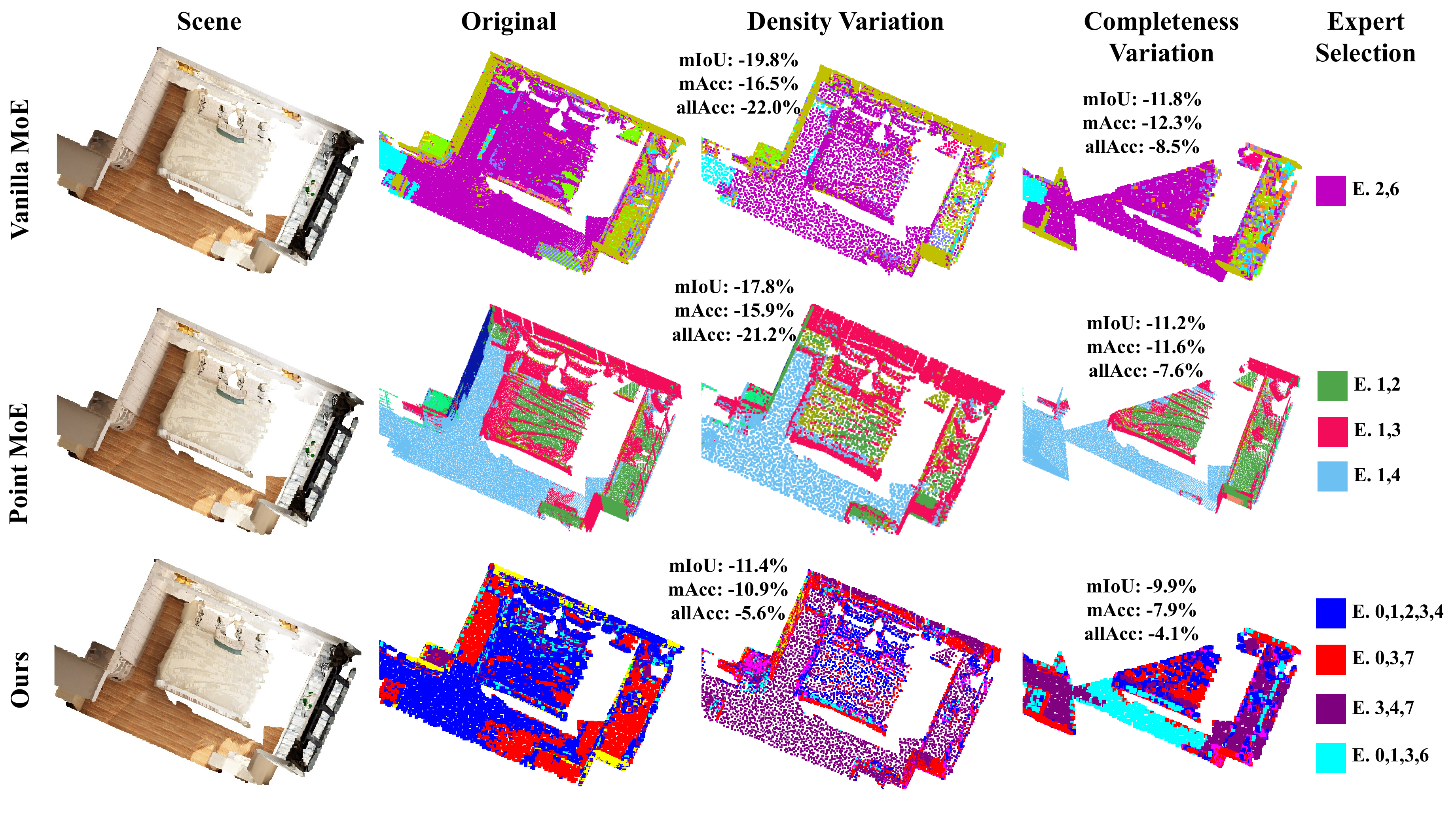}
    \caption{\textbf{Routing behavior under topology variations.} Vanilla MoE denotes our standard feature-only MoE baseline. We compare Vanilla MoE, Point-MoE~\cite{chen2025point}, and STAR under controlled density and completeness perturbations. For samples from the same semantic category (e.g., “bed” or “floor”), feature-only routers tend to keep similar expert assignments despite local geometric changes, causing larger performance drops (e.g., a 19.8\% mIoU drop under density variation for Vanilla MoE). In contrast, STAR incorporates spatial-topology cues into routing, adjusts expert subsets, and shows smaller degradation under fine-grained geometric variations.}
    \label{fig:teaser}
    \vspace{-3mm}
\end{figure*}

Toward this goal, leveraging large-scale multi-source 3D data for unified scene understanding has become a prominent research direction~\cite{ppt,wang2024one,chen2025point,sonata}. However, joint learning across heterogeneous 3D domains remains challenging because different sensing, acquisition, and generation pipelines impose fundamentally different sampling structures on the same underlying geometry. For example, LiDAR produces sparse, ray-wise measurements, whereas RGB-D fusion, multi-view reconstruction, and mesh-surface sampling generally yield denser and more continuous surface observations. Consequently, the same semantic object, such as a wall, may exhibit markedly different local structures across domains, ranging from sparse and discontinuous point patterns to dense and nearly complete surfaces. This semantic consistency yet topological heterogeneity makes it difficult for a unified model to reconcile conflicting geometric patterns during joint training, potentially degrading learned representations and causing negative transfer across domains.

To address this challenge, existing joint training methods primarily follow two paradigms: unified representation learning and modularized adaptation. Unified representation learning seeks a shared feature space for multi-source data~\cite{pointcontrast,pointmae,sonata}. However, such alignment may suppress sensor-specific geometric details when fitting heterogeneous cross-domain distributions, compromising precision. In contrast, modularized adaptation paradigms maintain flexibility by introducing specialized modules into a shared backbone, yet they are limited by insufficient modeling of 3D heterogeneity. Static strategies such as~\cite{ppt,wang2024one} can only learn globally fixed parameters and do not respond dynamically to drastic variations in density within 3D point clouds. Although dynamic strategies such as 3D Mixture of Experts (MoE)~\cite{chen2025point,uni3dmoe,xu2025limoe} introduce routing mechanisms, most routers use intermediate task features as routing inputs. Under semantic supervision, such feature-only routing may underrepresent local sampling topology, as illustrated in Figure~\ref{fig:teaser} and further discussed in Section~\ref{sec:experiments}. As a result, when semantically similar objects exhibit different density, completeness, or neighborhood patterns across sensors, expert allocation can become suboptimal, leading to degraded cross-domain performance.

Thus, we propose STAR, a topology-sensitive routing framework for multi-domain 3D scene understanding. Our primary setting is multi-domain joint training: STAR learns a shared backbone from multiple source datasets, while dataset-specific heads or fine-tuning are used only when label spaces or task formats differ. STAR has two complementary branches. The frozen Unified Representation branch (Re) uses multi-attribute self-supervised pre-training to provide stable cross-domain structural priors. The Domain-aware branch (Do) introduces Domain-Spatial-Guided Routing (DSR) and Entropy-controlled Dynamic Allocation (EDA) for adaptive expert allocation. DSR injects local spatial context into routing, enabling expert selection to respond to density, completeness, and neighborhood-structure variations, while EDA adjusts activated experts according to routing uncertainty for stable training. Through these branches, STAR combines stable cross-domain representation learning with topology-sensitive expert routing. As illustrated in Figure~\ref{fig:radar}, STAR shows consistent gains across multiple 3D understanding benchmarks. Our contributions are as follows:

\begin{itemize}[label=$\bullet$]
\item \textbf{Topology-sensitive routing for 3D MoE.} We identify that feature-only routers may underrepresent sensor-induced local geometric variations under semantic supervision. STAR incorporates spatial context into routing, improving sensitivity to density, completeness, and neighborhood-structure variations.
\item \textbf{Synergistic design of dual-branch framework.} STAR decouples representation learning into two mutually reinforcing branches. The frozen Re branch employs self-supervised augmentations simulating physical variations to filter domain noise and establish a robust cross-domain structural anchor. Grounded in this prior, the Do branch leverages DSR and EDA for highly adaptive, topology-aware expert allocation.

\item \textbf{Comprehensive evaluation.} Extensive experiments on multiple indoor and outdoor 3D understanding benchmarks demonstrate that STAR consistently outperforms existing approaches. Specifically, it achieves 80.1\% mIoU on ScanNet Val and 77.2\% mIoU on S3DIS, showing consistent gains over strong baselines.
\end{itemize}

\section{Related Work}
\subsection{3D Understanding}
3D scene understanding is fundamental to computer vision, covering tasks such as semantic segmentation~\cite{qi2017pointnet, qi2017pointnet++, thomas2019kpconv, pointcontrast, qian2022pointnext}, object detection~\cite{zhou2018voxelnet, lang2019pointpillars, shi2020pv}, and instance segmentation~\cite{jiang2020pointgroup, he2022pointinst3d}. While early voxel-based methods faced scalability limits, modern point-based~\cite{qi2017pointnet++, thomas2019kpconv} and transformer-based~\cite{ptv2, ptv3} architectures have significantly improved performance. However, most models remain domain-specific and lack robust cross-domain generalization. 
In this context, our proposed STAR serves as a generalizable 3D backbone that improves domain-adaptive representation learning for downstream 3D understanding tasks.

\subsection{Unified and Adaptive 3D Representation Learning}
Building a generalizable 3D scene understanding model requires balancing cross-domain generalization and domain-specific adaptation. Current joint training paradigms can be roughly categorized into two groups. Unified representation learning constructs a shared feature space through large-scale self-supervised pre-training~\cite{pointcontrast, pointmae,sonata, zhou2023uni3d,zhang2025concerto,kolodiazhnyi2025unidet3d,soum2023mdt3d}. While effective for generalization, such shared representations may suppress sensor-specific geometric details when aligning heterogeneous point-cloud distributions. In contrast, modularized adaptation paradigms~\cite{beyondsparse,ppt,wang2024one} introduce specialized parameters to improve flexibility, but are usually less responsive to fine-grained local topology variations. Different from purely unified or static adaptation strategies, STAR complements shared structural priors with topology-sensitive routing. By incorporating local geometric cues into expert allocation, STAR enables fine-grained adaptation across heterogeneous point clouds while preserving cross-domain representation stability.
\begin{figure*}[t]
    \centering
    \includegraphics[width=0.92\linewidth]{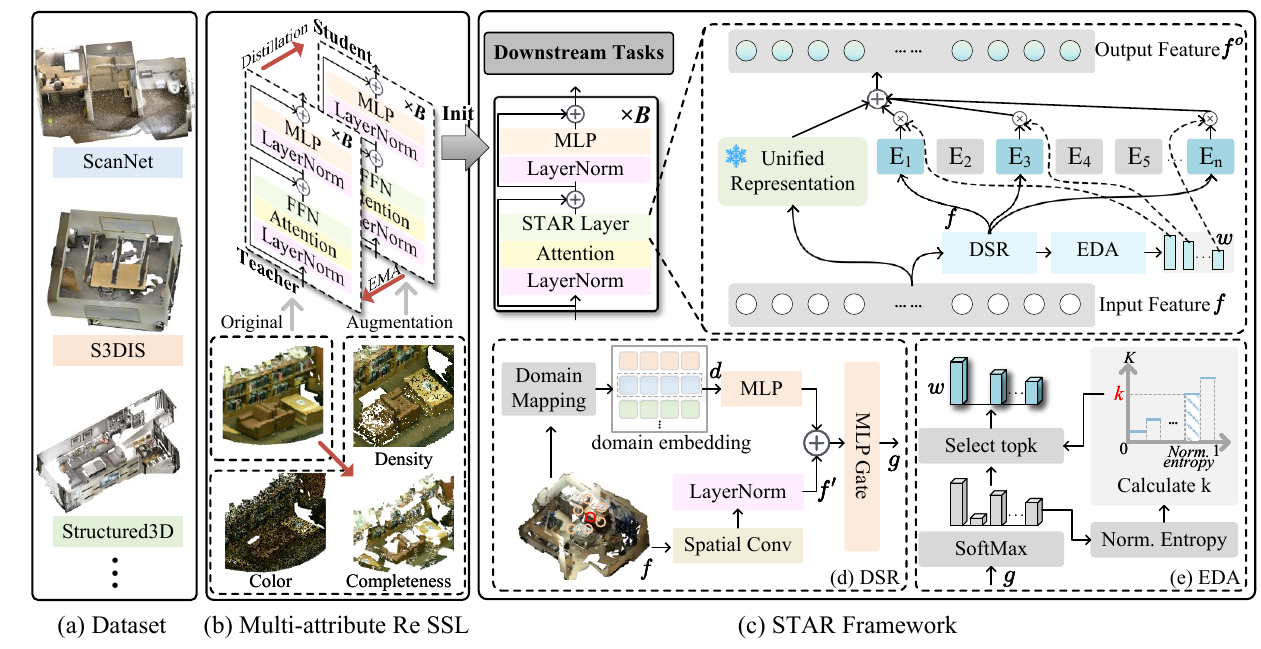}
    \caption{\textbf{STAR architecture.} It combines two branches: a domain-aware branch (Do) and a unified-representation branch (Re). STAR starts with a pretrained model generated through self-supervised learning on multi-attribute data. Re freezes this pretrained knowledge to extract cross-domain geometric and structural patterns. Do uses Domain-Spatial-Guided Routing (DSR) and Entropy-Controlled Dynamic Allocation (EDA) to capture domain-specific features. Together, these branches enhance both domain adaptability and generalization.}
    \label{fig:framework}
    \vspace{-4mm}
\end{figure*}
\subsection{Mixture-of-Experts}
Mixture of Experts (MoE) has been extensively applied to address multi-domain joint training and domain generalization. In 2D vision, existing works \cite{li2022sparse,dai2021generalizable,xu2024cbdmoe,zhong2022meta} integrate experts into Transformer architectures to achieve effective adaptation for large-scale image distributions. Concurrently, recent research has begun to explore the application of MoE in the 3D domain, including Point-MoE \cite{chen2025point}, Uni3D-MoE \cite{uni3dmoe}, and LiMoE \cite{xu2025limoe}. However, most existing 3D MoE methods route tokens using learned task features. While effective, these features may not explicitly encode local sampling topology, making expert allocation less responsive to density, completeness, and neighborhood-structure variations. In contrast, STAR introduces dynamic spatial routing to perceive such variations and adapt expert scheduling across heterogeneous sensor distributions.

\section{Method}

\subsection{Overview}
We propose STAR,  a spatial-topology aware routing framework for generalizable 3D understanding. It jointly models domain-aware expert features and unified representation features, enabling adaptive modeling of diverse domain distributions while maintaining cross-domain consistency and generalization capability. The overall architecture is illustrated in Figure \ref{fig:framework}. Specifically, we first employ a teacher-student network framework \cite{mean_teacher} to conduct multi-attribute self-supervised learning across multiple datasets, thereby learning geometric domain-structure priors that generalize across domains. Subsequently, the STAR architecture undergoes multi-domain supervised joint training, with its weights initialized from the student network. 
FFN weights are duplicated for both the Re and Do branches, while the Re branch remains frozen. The Do branch consists of two components: Domain-Spatial-Guided Routing (DSR) and Entropy-Controlled Dynamic Allocation (EDA). DSR leverages local geometric cues to enable the model to perceive topological variations, while EDA maintains the stability of the system by balancing the distribution of experts. By integrating these two mechanisms, the model adaptively activates expert subsets conditioned on spatial and source-domain structural cues.
\subsection{Unified Representation Branch}
We introduce a frozen Unified Representation branch (Re), constructed via large-scale multi-attribute self-supervised learning to provide robust cross-domain feature representations. Motivated by common discrepancies across point-cloud domains, we focus on three variation factors that affect cross-domain generalization: color distribution, point density, and object completeness caused by occlusions or limited viewpoints. Accordingly, we design three self-supervised alignment tasks tailored to color, density, and completeness, respectively. Specifically, we partition the point cloud into multiple patches. Within each patch, points are randomly assigned black coloration at varying ratios and probabilities, while certain points are also randomly discarded. Additionally, we apply masking operations to entire patches to simulate variations in point cloud completeness. Inspired by DINOv2 \cite{dinov2} and Sonata \cite{sonata}, we adopt a feature distillation framework based on a teacher-student network architecture, where the teacher's weights are updated via an exponential moving average (EMA) of the student's weights. The teacher receives raw data, while the student receives augmented data. The student is trained to align with the teacher through a cluster-based loss \cite{caron2020unsupervised,sablayrolles2018spreading}, ensuring uniform feature consistency across diverse augmentations of the same point cloud. This encourages invariance to common domain perturbations, providing stable features for domain-specific experts. Furthermore, we initialize the STAR architecture with the weights of the student network, duplicating its FFN for all experts in both Re and Do branches, while keeping Re branch frozen. Given point-level tokens $f \in \mathbb{R}^{N \times D}$, where $N$ denotes the number of tokens and $D$ denotes the feature dimension, the output of Re branch is: $f^{\text{Re}} = E_{Re}(f),$
where $E_{Re}$ denotes the expert in the Re branch.
\subsection{Domain-Spatial-Guided Routing}
To achieve topology-aware expert selection, we design DSR. 
Given $f$ and its corresponding domain embedding $d$, we first reshape ${f}$ into a 3D sparse tensor to capture its local topological structure and positional correlations in the spatial dimensions, and then apply a 3D spatial convolution operation for feature extraction. Through a set of learnable convolutional kernels and normalization, the model effectively extracts feature representations ${f}'$ with spatial locality awareness. 
Subsequently, we map the current scene to the corresponding domain embedding $d$ based on its dataset affiliation, and a lightweight MLP then transforms $d$ into a continuous vector $\mathbf{e}_d \in \mathbb{R}^D$ for channel alignment, which encodes source-domain structural priors. We add ${e}_d$ to the spatially convolved feature ${f}'$ via broadcasting to generate the domain-aware routing input: ${z} = f' + e_d,$
where $z \in \mathbb{R}^{N\times D}$. This fusion makes routing depend on both local spatial context and source-domain structural information, thereby enhancing the domain adaptability of expert assignment.
The resulting ${z}$ is then fed into a gating network $\mathcal{G}$, composed of a MLP with nonlinear activation functions. $\mathcal{G}$ outputs routing logits: ${g} = \mathcal{G}(z) \in \mathbb{R}^{ N \times K},$
where $K$ is the number of experts.
\subsection{Entropy-Controlled Dynamic Allocation}
To achieve robust and adaptive expert allocation, we propose EDA.
First, we apply the softmax function to the gating outputs along the last dimension to obtain a probability distribution $p = \text{SoftMax}(g) \in \mathbb{R}^{ N \times K}$, and calculate the Shannon entropy for each token:
\begin{equation}
    H = -\sum_{j=1}^K p[:,j] \odot \log p[:,j],
\end{equation}
where $\odot$ denotes element-wise multiplication and $H\in \mathbb{R}^{ N }$ reflects the model's decision-making uncertainty for that token.
Next, we linearly map the entropy values to the number of experts $k$ to dynamically determine the number of activated experts:

\begin{equation}
k = \left\lceil k_{\min} + \frac{H}{H_{\max}} \cdot (k_{\max} - k_{\min}) \right\rceil,
\end{equation}
where $k\in \mathbb{R}^{ N}$ represents the number of selected experts, $H_{\max} = \log K$ represents the theoretical maximum entropy, with $k_{\min}=1$ and $k_{\max}=K$, and $ \left\lceil \cdot \right\rceil$ denotes the ceiling function. 
Tokens with higher entropy (uncertainty) activate more experts to enhance representation capacity, whereas low-entropy tokens activate fewer experts to improve computational efficiency. 
We sort the experts in descending order of their probabilities $ p $, and activate the top-$ k $ experts with the highest probabilities. Each expert's weight assignment is given by:
\begin{equation}
w[i,j] =
\begin{cases}
p[i,j], & j \in E_i^{\text{act}} \\
0, & otherwise,
\end{cases}
\end{equation}
where $ E_i^{\text{act}} $ denotes the activated expert indices for the i-th token and $w \in \mathbb{R}^{N\times K}$.

\begin{table*}[t]
\small
\centering
\caption{\textbf{Indoor semantic segmentation.}}
\begin{tabular}{l *{10}{c}}
\toprule
\multirow{2}{*}{Method}&\multirow{2}{*}{Source}& \multicolumn{3}{c}{ScanNet Val} & \multicolumn{3}{c}{ScanNet200 Val} & \multicolumn{3}{c}{S3DIS Area 5} \\
\cmidrule(lr){3-5} \cmidrule(lr){6-8} \cmidrule(lr){9-11}
 && mIoU & mAcc & allAcc & mIoU & mAcc & allAcc & mIoU & mAcc & allAcc \\
\hline
PTv3 \cite{ptv3} & CVPR 2024& 77.6 & 85.0 & 92.0 & 35.3 & 46.0 & 83.4 & 73.4 & 78.9 & 91.7 \\

PPT \cite{ppt} & CVPR 2024& 78.6 & 85.9 & 92.3 & 36.0 & 46.2 & 83.8 & 74.3 & 80.1 & 92.0 \\
Point-MoE \cite{chen2025point} & ICLR 2026& 77.2 & 85.0 & 92.0 & 36.2 & 44.5 & 83.8 & 72.9 & 78.1 & 90.9 \\
Point-MoE+Re \cite{chen2025point} & ICLR 2026& 78.2 & 86.7 & 92.2 & 36.5 & 45.2 & 84.1 & 74.1 & 79.8 & 91.8 \\
Sonata \cite{sonata} & CVPR 2025& 79.4 & 86.1 & 92.5 & 36.8 & 46.5 & \textbf{84.4} & 76.0 & 81.6 & 93.0 \\
\rowcolor[HTML]{caeef0} STAR (Ours) & - & \textbf{80.1} & \textbf{87.3} & \textbf{93.1} & \textbf{37.2} & \textbf{47.9} & \textbf{84.4} & \textbf{77.2} & \textbf{82.5} & \textbf{93.1} \\
\bottomrule
\end{tabular}
\label{tab:results}
\end{table*}

We apply load-balancing loss \cite{switch_transformer} to prevent expert imbalance:
\begin{equation}
\begin{split}
    \mathcal{L}_{\text{balance}} &=  {K} \cdot \sum_{j=1}^K c_j \cdot  r_j ,\\
    c_j = \frac{1}{N} \sum_{i=1}^{N} 	
\mathds{1} \{j &\in E_i^{act}\}, 
    r_j = \frac{1}{N} \sum_{i=1}^{N} p[i,j],
\end{split}
\end{equation}
where $c_j$ is the proportion of tokens routed to expert $j$, and $r_j$ is the average probability assigned by DSR via the softmax function. This loss promotes uniform routing, fostering collaborative learning across all experts.

Finally, the Do branch output is computed as a weighted sum of the top-$k$ experts selected by routing probabilities. 
\begin{equation} \small    
    f^{\text{Do}} = \sum_{j=1}^{K} w[:, j] \odot E_j(f).
\end{equation}
The domain-aware branch accurately captures spatial contexts for targeted expert selection, ensuring stable, robust, and adaptive allocation across diverse domains. 
The output feature is computed as: $f^{o} = f^{\text{Do}} + f^{\text{Re}}.$

\subsection{Training Recipe}
We focus on multi-domain joint training for 3D scene understanding: a shared backbone is trained on source datasets, while dataset-specific heads or fine-tuning are used only for different label spaces or task formats. First, we obtain a pretrained model through multi-attribute self-supervised learning and use the student model's weights to initialize the parameters of the STAR network. 
Subsequently, we adopt multi-dataset joint training as in PPT~\cite{ppt}, unifying category representations through a CLIP-head and InfoNCE loss~\cite{infonce}. The training loss is:
\begin{equation}
    \mathcal{L}_{\text{joint}} = \mathcal{L}_{\text{InfoNCE}} + \lambda \mathcal{L}_{\text{balance}}.
\end{equation}

After completing the joint training, the resulting model can be directly deployed for the primary tasks within the joint training framework and further fine-tuned to address diverse downstream tasks on novel datasets. 

Taking semantic segmentation and multimodal detection as examples, the segmentation loss is defined as the standard cross-entropy loss:
\begin{equation}
\mathcal{L}_{\text{seg}} = -\frac{1}{M} \sum_{i=1}^{M} \sum_{c=1}^{C} y_{i,c} \log(q_{i,c}),
\end{equation}
where $M$ denotes the number of samples, $C$ is the number of classes, $y_{i,c}$ represents the ground-truth label, and $q_{i,c}$ is the predicted probability for class $c$ of sample $i$.

For the multi-modal detection task, we follow SpatialLM \cite{SpatialLM} and leverage the autoregressive property of the Qwen2.5 language model \cite{qwen2} to treat coordinate prediction as a sequence generation task. We adopt the same standard cross-entropy loss as used in Qwen \cite{qwen2025qwen25technicalreport}:
\begin{equation}
\mathcal{L}_{\text{det}} = -\frac{1}{T} \sum_{t=1}^{T} \log P(w_t | w_{<t}, \theta),
\end{equation}
where $T$ is the length of the coordinate sequence, $w_t$ denotes the $t$-th coordinate element in the sequence, $w_{<t}$ represents the history of previously generated coordinates, and $\theta$ denotes the model parameters.



\begin{figure*}[t]
    \centering
    \includegraphics[width=0.95\linewidth]{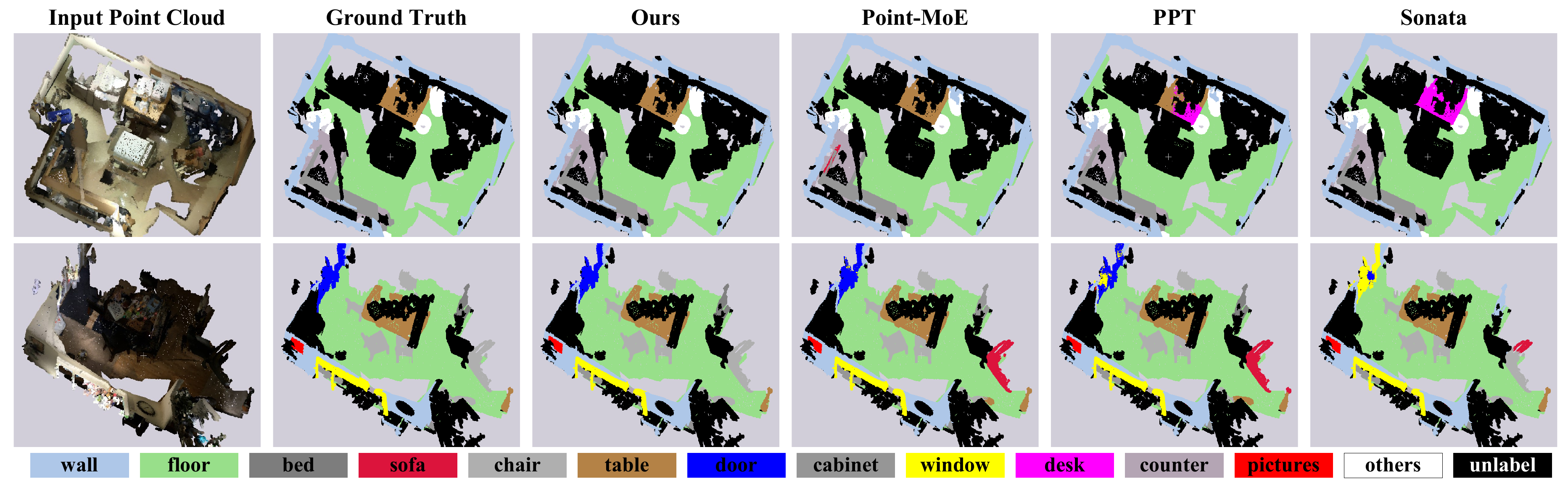}
    \caption{\textbf{Qualitative analysis.} Visualization of different methods on ScanNet.}
    \label{fig:visualizaiton}
\end{figure*}
The overall fine-tune training objective is:

\begin{equation}
    \mathcal{L}_{\text{ft}} = \mathcal{L}_{\text{task}} + \lambda \mathcal{L}_{\text{balance}},
\end{equation}
where $\mathcal{L}_{\text{task}}$ denotes the task-specific loss (e.g., $\mathcal{L}_{\text{seg}}$ for segmentation, $\mathcal{L}_{\text{det}}$ for detection).

\section{Experiments}
\label{sec:experiments}
\subsection{Implementation Details}
We conduct self-supervised pretraining on six datasets: ScanNet~\cite{dai2017scannet}, S3DIS~\cite{s3dis}, Structured3D~\cite{zheng2020structured3d}, 3D-Front~\cite{3dfront}, ARKitScenes~\cite{baruch2021arkitscenes}, and HM3D~\cite{hm3d}, totaling 47,273 training samples. The network follows Sonata, using 5 stages with block counts of 3, 3, 3, 12, and 3 per stage. Pretraining uses a batch size of 64, a learning rate of 0.0004, and 50 epochs. Following Sonata~\cite{sonata}, patch masking employs a cosine scheduler. Density variations and color dropout use different sampling ratios across patches. Further details are provided in the supplementary material. For STAR, the maximum expert number $K$ is 8 and $\lambda$ is 0.001. The encoder architecture closely follows the pretrained network, with Re and Do branches added only to the final block of each stage. Domain embeddings are randomly initialized. All experiments are conducted on 8 NVIDIA A100 GPUs, with the AdamW optimizer employed throughout.
\begin{table}[t]
\small
\centering
\setlength{\tabcolsep}{4pt} 
\caption{\textbf{Outdoor semantic segmentation.} $^\star$ indicates results reproduced from the official repository.}
\begin{tabular}{l|ccc ccc}
\toprule
\multirow{2}{*}{Method} & \multicolumn{3}{c}{nuScenes Val} & \multicolumn{3}{c}{Waymo Val} \\
\cline{2-4} \cline{5-7}
 & mIoU & mAcc & allAcc & mIoU & mAcc & allAcc \\
\hline
PTv3 \cite{ptv3}& 80.4 & 87.2 & 94.7 & 71.3 & 80.5 & 94.7 \\
Sonata$^\star$ \cite{sonata} & 81.2 & \textbf{87.7} & \textbf{94.8} & 72.1 & 82.6 & \textbf{94.8} \\
\rowcolor[HTML]{caeef0}STAR (Ours) & \textbf{81.7} & 87.4 & \textbf{94.8} & \textbf{72.7} & \textbf{82.7} & \textbf{94.8} \\
\bottomrule
\end{tabular}
\label{tab:outdoor_result}
\vspace{-4mm}
\end{table}
\subsection{Experimental Results}
\noindent \textbf{Indoor Semantic Segmentation.}
Following PPT \cite{ppt}, we conduct joint training on three datasets: ScanNet (20 classes), S3DIS (13 classes), and Structured3D (25 classes). To achieve cross-dataset semantic alignment of categories, we incorporate a CLIP-based classification head. We directly evaluate performance on ScanNet and S3DIS. Furthermore, we fine-tune our model on the more challenging ScanNet200 benchmark, achieving strong performance across all three core metrics (mIoU, mAcc, and allAcc), as detailed in Table \ref{tab:results}. Specifically, our method attains mIoU scores of 80.1\%, 37.2\%, and 77.2\% respectively, representing improvements of 0.5\%, 0.4\%, and 1.2\% over Sonata. 
Figure \ref{fig:visualizaiton} presents a qualitative comparison with other methods. In the first scenario, given an incomplete chair, other approaches misidentify it as a sofa, whereas ours recognizes it as a chair. This demonstrates that the Re obtained through SSL significantly enhances the model's capability to extract discriminative features from incomplete objects. In the second scenario, while other methods erroneously classify a table as a desk, our approach leverages DSR to integrate surrounding spatial context. This helps route inputs to suitable experts through spatial context, improving recognition of ambiguous objects. These results suggest better cross-domain adaptability in complex indoor scenarios.

\noindent \textbf{Outdoor Semantic Segmentation. }
Following the established indoor experimental setup, we conduct comprehensive joint training and direct cross-dataset evaluation on the nuScenes \cite{caesar2020nuscenes} and Waymo \cite{waymo} benchmarks. As presented in Table \ref{tab:outdoor_result}, our model achieves mIoU scores of 81.7\% and 72.7\% on the nuScenes and Waymo datasets respectively, outperforming Sonata by 0.5\% and 0.6\%. These quantitative results validate STAR's improved generalization capability and robustness in challenging outdoor scenarios.

\noindent \textbf{Extension to Unseen Scenes. }
To validate STAR's zero-shot generalization to unseen domains, we evaluate on two datasets—SpatialLM and Matterport3D—using three different domain embeddings for inference, as shown in Table~\ref{tab:zero_shot_unseen}.
SpatialLM is derived from professional CAD interior models, with point clouds generated by sampling virtual meshes. The data contains no real-world noise, and structural surfaces such as walls and floors form complete, continuous topologies. Using the Structured3D domain embedding yields the best performance (mIoU = 38.7), clearly outperforming Sonata (36.0) and Point-MoE (36.4). This result is consistent with the data source: Structured3D is also derived from synthetic sampling of CAD interior models, allowing DSR to use similar geometric characteristics for expert allocation.
Matterport3D \cite{chang2017matterport3d} is a large-scale complex indoor scene dataset captured by real RGB-D sensors, featuring substantial real-world noise, occlusion gaps, and non-uniform point density. Using the ScanNet domain embedding achieves the best performance (mIoU = 49.5), surpassing Sonata (48.1) and Point-MoE (41.8). This is because ScanNet is also a real RGB-D indoor scan dataset, sharing similar sensor patterns, noise characteristics, and geometric distributions with Matterport3D, which helps route features toward experts adapted to real scan data.
These experiments suggest that, for an unseen domain, STAR can select a source-domain embedding according to available acquisition metadata or sampling characteristics, enabling zero-shot transfer without target-domain training.

\begin{table}[t]
    \small
    \centering
    \setlength{\tabcolsep}{0.5pt}
    \caption{\textbf{Zero-shot generalization to unseen scenes.} Results on SpatialLM and Matterport3D val using different domain embeddings for inference.}
    \label{tab:zero_shot_unseen}
    \begin{tabular}{lcccccc}
        \toprule
        \multirow{2}{*}{Method} & \multicolumn{3}{c}{SpatialLM} & \multicolumn{3}{c}{Matterport3D} \\
        \cmidrule(lr){2-4}\cmidrule(lr){5-7}
        & mIoU & mAcc & allAcc & mIoU & mAcc & allAcc \\
        \midrule
        Point-MoE~\cite{chen2025point} & 36.4 & 42.6 & 68.2 & 41.8 & -- & -- \\
        Sonata~\cite{sonata}           & 36.0 & 43.6 & 68.6 & 48.1 & 61.0 & 77.6 \\
        \midrule
        STAR (ScanNet emb.)      & 35.5 & 41.7 & 67.8 & \textbf{49.5} & \textbf{62.1} & \textbf{78.6} \\
        STAR (S3DIS emb.)        & 32.5 & 39.6 & 69.3 & 47.8 & 60.3 & 77.8 \\
        STAR (Structured3D emb.) & \textbf{38.7} & \textbf{46.5} & \textbf{72.0} & 47.7 & 61.3 & 77.6 \\
        \bottomrule
    \end{tabular}
    \vspace{-3mm}
\end{table}

\begin{table}[t]
\small
\centering
\caption{\textbf{Multimodal object detection} results on the ARKitScenes validation set.}

\begin{tabular}{l|cc}
\toprule
Method & F1@0.25 & F1@0.5 \\
\hline
SpatialLM \cite{SpatialLM} + Sonata \cite{sonata}& 58.9 & 49.5 \\
\rowcolor[HTML]{caeef0} SpatialLM \cite{SpatialLM} + STAR (Ours) & \textbf{60.8} & \textbf{51.9} \\
\bottomrule
\end{tabular}
\label{tab:detect_comparison}
\vspace{-3mm}
\end{table}

\noindent \textbf{Multimodal Object Detection. }
Beyond segmentation, our method also extends effectively to detection tasks. In this experiment, we implement the SpatialLM \cite{SpatialLM} for object detection on the ARKitScenes dataset \cite{baruch2021arkitscenes}. 
Evaluation employs F1-score metrics under two IoU thresholds (0.25 and 0.5). We establish the baseline using officially fine-tuned SpatialLM weights with Sonata-based point cloud encoder, which serves as the initialization for STAR's subsequent fine-tuning. The optimization employs a learning rate of \(\mathbf{5\times10^{-5}}\) over 10 epochs. As shown in Table \ref{tab:detect_comparison}, our method achieves 60.8\% F1@0.25 and 51.9\% F1@0.5, outperforming the baseline by 1.9\% and 2.4\%. These results validate the cross-task effectiveness of our framework and demonstrate its potential for extension to diverse vision tasks.
\subsection{Ablation Studies}






\begin{table}[t]
\small
    \centering
    \caption{\textbf{Ablation study on each component.}}
    \label{tab:ablation}
    \resizebox{\columnwidth}{!}{
        \begin{tabular}{ccc|cccccc}
            \toprule
            \multirow{2}{*}{Re} & \multirow{2}{*}{DSR} & \multirow{2}{*}{EDA} & \multicolumn{3}{c}{ScanNet Val} & \multicolumn{3}{c}{S3DIS Val} \\
            \cline{4-6} \cline{7-9}
             & & & mIoU & mAcc & allAcc & mIoU & mAcc & allAcc \\
            \hline
            × & × & × & 77.5 & 85.4 & 92.1 & 73.5 & 78.9 & 92.0 \\
            $\checkmark$ & × & × & 78.8 & 85.9 & 92.5 & 75.7 & 81.4 & 92.5 \\
            $\checkmark$ & $\checkmark$ & × & 79.5 & 87.0 & 92.9 & 76.4 & 82.1 & 93.0 \\
            \rowcolor[HTML]{caeef0} $\checkmark$ & $\checkmark$ & $\checkmark$ & 80.1 & 87.3 & 93.1 & 77.2 & 82.5 & 93.1 \\
            \bottomrule
        \end{tabular}
    }
    \vspace{-3mm}
\end{table}

\begin{table*}[t]
    \small
    \centering
    \caption{Quantitative analysis of performance under two point cloud topological variations.}
    \renewcommand{\arraystretch}{1.0}
    \setlength{\tabcolsep}{3pt}
    \resizebox{\textwidth}{!}{%
    \begin{tabular}{l ccc ccc ccc ccc ccc ccc}
        \toprule
        \multirow{2}{*}{\textbf{Method}}
            & \multicolumn{3}{c}{\textbf{Original}}
            & \multicolumn{3}{c}{\textbf{Dropout 0.9}}
            & \multicolumn{3}{c}{\textbf{Dropout 0.7}}
            & \multicolumn{3}{c}{\shortstack{\textbf{mask\_size=0.6} \\ \textbf{mask\_ratio=0.5}}}
            & \multicolumn{3}{c}{\shortstack{\textbf{mask\_size=0.8} \\ \textbf{mask\_ratio=0.5}}}
            & \multicolumn{3}{c}{\shortstack{\textbf{mask\_size=0.8} \\ \textbf{mask\_ratio=0.8}}} \\
        \cmidrule(lr){2-4} \cmidrule(lr){5-7} \cmidrule(lr){8-10}
        \cmidrule(lr){11-13} \cmidrule(lr){14-16} \cmidrule(lr){17-19}
            & \textbf{mIoU} & \textbf{mAcc} & \textbf{allAcc}
            & \textbf{mIoU} & \textbf{mAcc} & \textbf{allAcc}
            & \textbf{mIoU} & \textbf{mAcc} & \textbf{allAcc}
            & \textbf{mIoU} & \textbf{mAcc} & \textbf{allAcc}
            & \textbf{mIoU} & \textbf{mAcc} & \textbf{allAcc}
            & \textbf{mIoU} & \textbf{mAcc} & \textbf{allAcc} \\
        \midrule

        Vanilla MoE
            & 78.5 & 85.8 & 92.3
            & 68.7 & 75.9 & 87.9
            & 75.2 & 82.4 & 91.0
            & 70.7 & 79.4 & 88.5
            & 66.8 & 76.4 & 87.3
            & 59.7 & 70.0 & 83.9 \\
        \rowcolor{gray!15}
        \quad $\Delta$
            & {---}  & {---}  & {---}
            & -9.8  & -9.9  & -4.4
            & -3.3  & -3.4  & -1.3
            & -7.8  & -6.4  & -3.8
            & -11.7 & -9.4  & -5.0
            & -18.8 & -15.8 & -8.4 \\
        \midrule
        Point-MoE
            & 77.2 & 85.0 & 92.0
            & 69.4 & 77.3 & 88.9
            & 73.9 & 81.7 & 90.8
            & 67.0 & 75.8 & 87.8
            & 67.7 & 76.6 & 88.2
            & 55.4 & 64.8 & 82.5 \\
        \rowcolor{gray!15}
        \quad $\Delta$
            & {---}  & {---}  & {---}
            & -7.8  & -7.7  & -3.1
            & -3.3  & -3.3  & -1.2
            & -9.5  & -8.4  & -3.8
            & -10.2 & -9.2  & -4.2
            & -21.8 & -20.2 & -9.5 \\
        \midrule
        
        Ours
            & 80.1 & 87.3 & 93.1
            & 74.1 & 81.5 & 90.7
            & 77.7 & 85.1 & 92.1
            & 73.9 & 82.5 & 90.4
            & 71.9 & 81.4 & 89.8
            & 63.2 & 73.4 & 86.0 \\
        \rowcolor{gray!15}
        \quad $\Delta$
            & {---}  & {---}  & {---}
            & -6.0  & -5.8  & -2.4
            & -2.4  & -2.2  & -1.0
            & -6.2  & -4.8  & -2.7
            & -8.2  & -5.9  & -3.3
            & -16.9 & -13.9 & -7.1 \\
        \bottomrule
    \end{tabular}%
    }
    \label{tab:degeneration_comparison}
\end{table*}

\noindent \textbf{Component Ablation. }
As shown in Table \ref{tab:ablation}, we validate STAR components on ScanNet and S3DIS.
Adding Re improves mIoU by 1.3\% and 2.2\% over the framework without STAR components, showing the benefit of shared pretrained representations.
Adding DSR with two experts brings another 0.7\% gain on both datasets, confirming the value of spatial context in routing. A DSR decomposition in the supplementary material shows that the spatial-convolution-only variant outperforms the domain-embedding-only variant (79.4\% vs. 79.1\% mIoU on ScanNet), while combining both reaches 79.5\%. This suggests that local spatial topology contributes more than domain embedding, with source-domain priors providing complementary guidance.
Finally, EDA improves performance to 80.1\% and 77.2\% mIoU by stabilizing expert utilization and activation counts. Overall, Re provides robust cross-domain representations, while DSR and EDA enable adaptive expert routing under complex distribution shifts.

\noindent \textbf{Efficiency Comparison. }
We analyze the efficiency–accuracy trade-off of STAR against previous methods.
Table~\ref{tab:efficiency} shows Sonata~\cite{sonata} outperforms PPT \cite{ppt} but at the cost of a large activated parameter increase.
In contrast, STAR further improves mIoU over Sonata (+0.7\%) with moderate parameter growth.
To verify that performance gains are not merely from model scaling, we implement two larger variants of Sonata.
The first (\#4) sets stage depths to [2, 2, 6, 2] and expands channels to [64, 128, 256, 512], while the second (\#5) increases all four decoder stage depths to 5.
Both variants have comparable parameters to STAR and follow official training settings.
Despite similar capacity, they still underperform STAR, demonstrating that our improvements primarily stem from the proposed routing design and dual-branch representation rather than parameter scaling.
The total per-sample latency is approximately 207.9ms, where DSR accounts for 7.9ms (3.8\%) and EDA accounts for 2.2ms (1.06\%), showing a minor impact on inference efficiency.
Overall, STAR provides a favorable accuracy–efficiency trade-off, converting limited additional computation into consistent improvements.
\begin{table}[t]
    \centering
    \caption{\textbf{Efficiency comparison.} ${'}$ and ${''}$ denote two scaled variants of Sonata. $^\dagger$ represents the Point-MoE reproduced using PTv3-L. ``Act. Params'' indicates the number of activated parameters. FPS is measured on one A100 GPU.}
    \label{tab:efficiency}
    \begin{tabular}{c|l|ccc}
    \toprule
    & Method & Act. Params & FPS & mIoU\\
    \hline
    \#1 &PTv3  \cite{ptv3} & 46.2M  & 11.0 & 77.6 \\
    \#2 &PPT  \cite{ppt} & 46.3M  & 7.2 & 78.6\\
    \#3 &Sonata \cite{sonata}& 124.8M & 5.9 & 79.4 \\
    \#4 &Sonata${'}$ & 147.6M & 5.3 & 79.4 \\
    \#5 &Sonata${''}$ & 148.6M & 4.9 & 79.6 \\
    \#6 &Point-MoE$^\dagger$ & 147.7M & 5.3 & 77.2 \\
    \rowcolor[HTML]{caeef0} \#7 & STAR (Ours)   & 147.5M & 4.9  & 80.1\\
    \bottomrule
    \end{tabular}
    \vspace{-3mm}
\end{table}

\begin{figure}[t]
    \centering
    \includegraphics[width=\columnwidth]{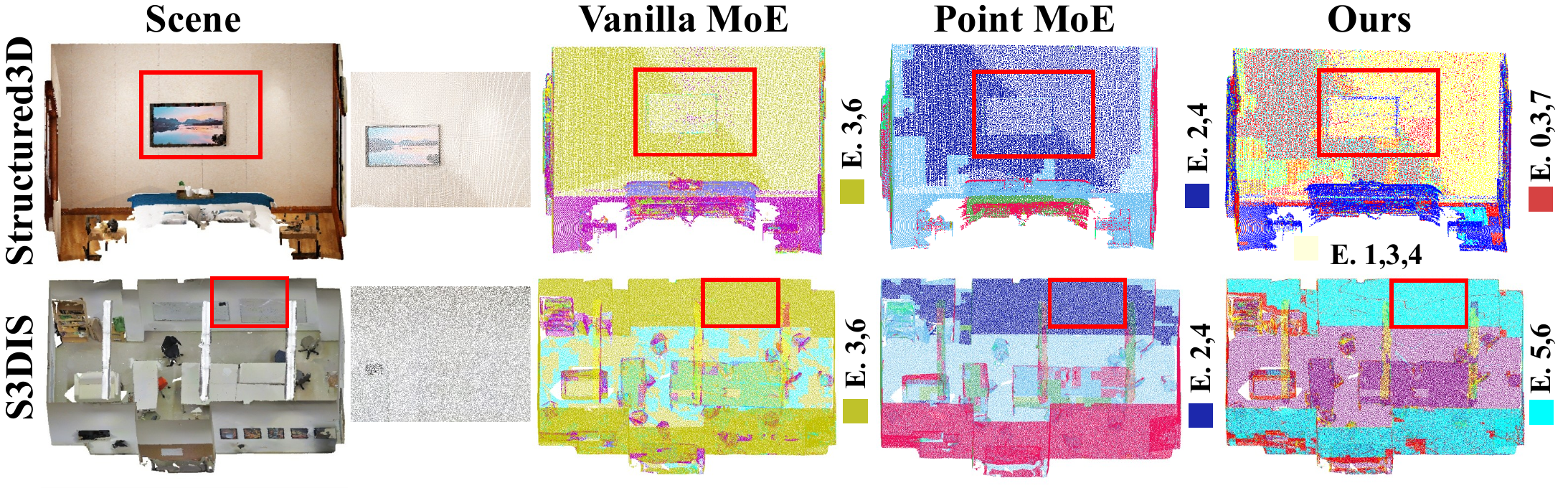}
    \caption{Expert activation across real cross-domain sampling differences. For the same ``wall'' category, S3DIS and Structured3D exhibit different local sampling statistics: 6,237 pts/m$^2$ / 0.91 cm vs. 5,512 pts/m$^2$ / 1.20 cm in density / 5-NN distance. STAR activates different expert subsets under these differences, while Vanilla MoE and Point-MoE produce more similar assignments.}
    \label{fig:fig6}
    \vspace{-5mm}
\end{figure}

\noindent \textbf{Expert Activation Analysis. }
As discussed in Section~\ref{sec:intro} and demonstrated in Figure~\ref{fig:teaser}, Vanilla MoE and Point-MoE~\cite{chen2025point} perform expert assignment primarily based on intermediate task features. Under semantic supervision, such routing may be less sensitive to local sampling topology. Therefore, when samples from the same semantic categories (e.g., ``bed'' or ``floor'') undergo local geometric fluctuations caused by occlusion or scanning patterns, Vanilla MoE tends to keep similar expert assignments, whereas STAR adjusts expert subsets according to spatial-topology cues.
Taking the ``wall'' category as an example, S3DIS and Structured3D exhibit different local sampling statistics (6,237 pts/m$^2$ / 0.91 cm vs. 5,512 pts/m$^2$ / 1.20 cm in density / 5-NN distance). As illustrated in Figure~\ref{fig:fig6}, STAR activates different expert subsets under these genuine cross-domain sampling differences, while Vanilla MoE and Point-MoE produce more similar assignments. This indicates that STAR responds to real cross-domain topological variations rather than merely to synthetic perturbation scenarios.

\noindent \textbf{Robustness Analysis Against Point Cloud Topological Variations.}
To evaluate STAR's robustness to point cloud topological variations, we conduct experiments on the ScanNet validation set and compare its performance with Vanilla MoE and Point-MoE~\cite{chen2025point}. We construct two representative perturbation scenarios: random dropout and regional masking. Random dropout stochastically removes a specified proportion of points to alter spatial density, whereas regional masking varies mask size and probability to simulate local structural loss caused by occlusions or sensor limitations in real-world scans. Table~\ref{tab:degeneration_comparison} shows STAR outperforms both baselines under standard inputs. Though topological perturbations affect STAR, it shows smaller performance drops and maintains consistent gains even in severe scenarios. This robustness primarily stems from DSR's topology-aware routing, which captures fine-grained local geometric context and dynamically adapts expert selection to point cloud structural changes instead of relying solely on high-level task features. EDA stabilizes routing distributions under perturbed inputs, preventing expert collapse and ensuring balanced activation. Detailed comparisons of expert activation groups across point cloud topological variations are provided in the supplementary material.


\section{Conclusion}
This paper presents STAR, a spatial-topology-aware routing framework for multi-domain 3D scene understanding. STAR combines a frozen Unified Representation branch for cross-domain structural priors with a Domain-aware branch using DSR and EDA for topology-sensitive expert allocation. Experiments on indoor and outdoor benchmarks show improved robustness and generalization under density and completeness variations.

{
    \small
    \bibliographystyle{ieeenat_fullname}
    \bibliography{main}
}

\end{document}